\documentclass[twoside]{IEEEtran}
\usepackage{xcolor}
\usepackage{amsmath}
\usepackage[utf8]{inputenc}
\usepackage{graphicx}
\usepackage{tabularx}
\usepackage{makecell}
\usepackage[english]{babel}
\usepackage{booktabs}
\usepackage{ragged2e}
\usepackage{multirow, balance}
\usepackage[noadjust]{cite}

\usepackage[section]{placeins}
\title{Reinforcement Learning for IoT Security: \\ A Comprehensive Survey}
\author{Aashma Uprety and Danda B. Rawat, \textit{Senior Member, IEEE}
\thanks{Manuscript received Day Month Year.} 
\thanks{Authors are with the Department of Electrical Engineering and Computer Science at Howard University, Washington, DC 20059, USA. E-mail: db.rawat@ieee.org.} 

\thanks{
This work was supported in part by the US NSF under grants CNS/SaTC 2039583, CNS 1650831 and 1828811, by the DoD Center of Excellence in AI and Machine Learning (CoE-AIML) at Howard University under Contract Number W911NF-20-2-0277 with the US Army Research Laboratory, the DoE's National Nuclear Security Administration (NNSA) Award \# DE-NA0003946. and the US Department of Homeland Security (DHS) under grant award number, 2017‐ST‐062‐000003. However, any opinion, finding, and conclusions or recommendations expressed in this material are those of the author and do not necessarily reflect the views of these funding agencies.
}}
{}

\begin{document}
\maketitle
\begin{abstract}
The number of connected smart devices has been increasing exponentially for different Internet-of-Things (IoT) applications. Security has been a long run challenge in the IoT systems which has many attack vectors, security flaws and vulnerabilities. Securing billions of  connected devices in IoT is a must task to realize the full potential of IoT applications. Recently, researchers have proposed many security solutions for IoT. Machine learning has been proposed as one of the emerging solutions for IoT security and Reinforcement learning is gaining more popularity for securing IoT systems. Reinforcement learning, unlike other machine learning techniques, can learn the environment by having minimum information about the parameters to be learned. It solves the optimization problem by interacting with the environment adapting the parameters on the fly. In this paper, we present an comprehensive survey of different types of cyber-attacks against different IoT  systems and then we present reinforcement learning and deep reinforcement learning based security solutions to combat those different types of attacks in different IoT systems. 
Furthermore, we present the Reinforcement learning for securing CPS systems (i.e., IoT with feedback and control) such as smart grid and smart transportation system.
The recent important attacks and countermeasures using reinforcement learning  in IoT are also summarized in the form of tables. With this paper, readers can have a more thorough understanding of IoT security attacks and countermeasures using Reinforcement Learning, as well as research trends in this area.
	
\end{abstract}
felix2020sur

\section{INTRODUCTION}
Internet of Things (IoT) connects the physical world to the digital world. It is a revolutionary technology in which machines talk to other machines to solve trivial to complex tasks \cite{atzori2010internet,felix2020sur, rawat2018performance, rawat2017vehicular}. Sensors and actuators are the resources from which data is exchanged between the physical world and the digital world. The sensors collect data that are to be stored and processed to provide service to the user. It has brought a drastic change in the lifestyle of humans by bringing smartness to the devices and will eventually increase the quality of human life. IoT has tremendously increased the use of the internet by bringing all the physical devices together in the network. Any physical device brought to internet connection that can interact with human can be an IoT device. For example, when cars are connected to each other through the internet and communicate with each other, this is called internet of cars.

IoT collects and processes human day to day data and brings automation to the task. With all the easiness provided by IoT, there also exist some pitfalls in using IoT. The major challenge is securing the system from attackers, maintaining the privacy of the user of IoT and making sure that certain IoT devices can be trusted. More the number of connected devices, more is the chance of the vulnerabilities to attack. Security in IoT operation is the major challenge to be faced by IoT designers. The dynamic environment of IoT and runtime communication adds additional security requirements on the IoT design. IoT brings flexibility and intelligence to the devices providing us usability but at the same time, it is also fearsome to use it. IoT is gaining a status for insecurity. Researchers divulge the dangerous flaws in IoT which poses a major challenge in IoT success \cite{mansfield2018open, min2018learning, felix2020sur}. We are sharing our every personal information through IoT devices and it is very important that our data are confidential.

Reinforcement learning is a machine learning approach in which the agent interacts with the environment and tries to maximize the numerical reward \cite{sutton2018reinforcement}. Human brain interacts with the outer environment and uses that interaction to understand and sustain in that environment \cite{5227780}. Reinforcement Learning uses the human brain and sensory processing system \cite{schultz1997neural} as an analogy to learning the environment. It is a process in which an agent has to explore all the system to understand it. Considering the time it takes to converge and get an optimal policy, it is not feasible in many scenarios. Traditional RL suffers a curse of dimensionality. As the environment becomes complex, there is exponential growth in the parameters to be learned by RL agent \cite{barto2003recent}. As a solution, we have deep reinforcement learning (DRL) which is a combination of deep network and reinforcement learning (RL) \cite{hinton2006reducing}. RL has been applied in securing IoT technology in various domains which is the main scope of this paper. IoT is a highly mobile technology and is very vulnerable to many cyber attacks. The sensors and actuators are one point of attack. Network for communication is again another major point of attacks in IoT. Much research work has been done to provide security to IoT system using RL technology.

The main scope of this paper is to provide a literature review of research done on securing IoT devices using RL. Along with this, we also provide a background of reinforcement learning. The paper is organized as follows. In Section II, we compare reinforcement learning with other machine learning techniques. Further, we discuss why RL is suitable for IoT scenarios. Section III is about research works related to RL on securing IoT from several threats. We present the application of RL in specific CPS system in Section IV. At the end some discussion and future research challenge is presented. Table \ref{my-label1} is the list of abbreviations we used throughout the paper.

\section{Brief Overview of Reinforcement Learning}
In this section, we briefly introduce reinforcement learning and talk about deep reinforcement learning. The comparison of RL and other machine learning techniques is presented. In the end, we answer why the use of RL is effective in the IoT scenario.

\subsection{Reinforcement Learning}
\begin{figure}[ht!]
	\centering
	\includegraphics[width=80mm]{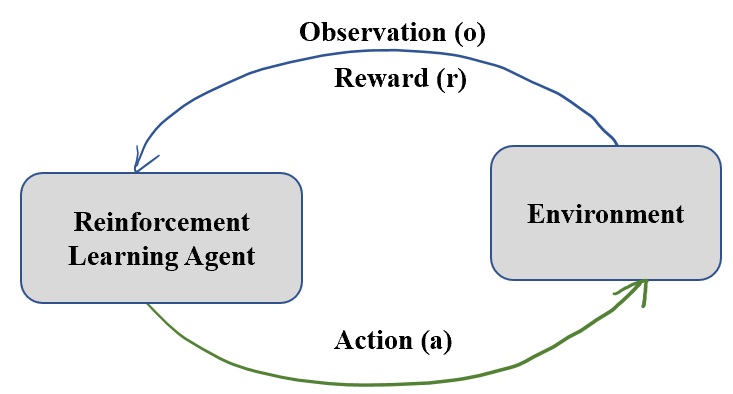}
	\caption{Agent-Environment Interaction in Reinforcement Learning.}
	\label{rlstate}
\end{figure}

Reinforcement learning is a kind of machine learning in which AI agent aims to accomplish a task by taking the best next step which can give them overall higher final reward, as shown in Fig. \ref{rlstate}.
In RL setting, the agent goes through many trial and error steps and tries to maximize the reward it gets from the environment \cite{sutton2018reinforcement}.  An agent interacts with an environment, which can be a simulator, a game, the real world etc. Each time step the agent observes the state $s_t$ from the environment, selects an action $a_t$, and then receives a reward $r_t$ and the environment changes to $s_{t+1}$. Therefore, each time step the agent gathers experiences ($s_t$, $a_t$, $r_t$, $s_{t+1}$) from which it can learn. If the action taken was favorable for the given environment, it will get a positive reward. If not it gets a negative reward. The agent continues to collect the reward aiming to maximize expected return from each state \cite{mnih2016asynchronous}. Reinforcement learning is a Markov Decision Process (MDP) in which the output of taking an action from a state depends only on the present state irrespective of past states and actions. MDP is a tuple consisting of five elements as
($S$,$A$,$P$,$R$,$\gamma$). It uses discount factor $\gamma$ which is a scalar value between 0 and 1. The discount factor is considered to maximize the future rewards that the agent gets from the environment. Value function in RL is a mapping from states to real numbers, where the value of a state represents the long-term reward achieved starting from that state and executing a particular policy. Value function $v(s)$ is a representation of how good it is for an agent to be in the state $s$. Bellman equation\cite{dreyfus2002richard} is the foundation mathematics behind reinforcement learning.
\[v(s)= E[R_{t+1} + \gamma v(S_{t+1}) | S_t= s]\]
In the given Bellman equation, the value function is decomposed as an immediate reward plus the value at the next successor state with discount factor($\gamma$).

\subsection{Deep Reinforcement Learning}
Deep Reinforcement Learning (DRL) is a combination of deep learning and RL. DRL is revolutionary research in RL which is capable to solve complex computational tasks \cite{franccois2018introduction}. For the complex environment, an approximation of value function and policy gradient is a complex task. For this,  deep network is used to approximate these values. Consider the set of actions taken by agents that results in a positive reward. In this case, a normal gradient is used to increase the probability of again taking these sets of actions. The deep network adds intelligence to RL agents and hence it accelerates the agent's capability to optimize the policy. RL is the only machine learning technique that can learn without any dataset. However, as the agent interacts with the environment, it generates the dataset. These datasets are used to train the deep network in DRL. Researchers have proposed many DRL approaches with its application ranging from control \cite{7983780}, resource management \cite{mao2016resource, zhang2017intelligent}, robotics \cite{vecerik2017leveraging, gu2017deep} and many more.
In 2015, Google DeepMind introduced deep Q-network (DQN)  \cite{mnih2015human}, delivering results exceeding human in playing Atari games. Deep neural
network was used in DQN as the function approximator. In Go games, AlphaGo \cite{silver2016mastering} and AlphaGo Zero \cite{lillicrap2015continuous} also showed an excellent result. Following that, DeepMind team made additional improvements based on DQN which builds a target DQN which calculates the maximum Q value and they named it Double DQN \cite{van2016deep}. Dueling DQN \cite{wang2015dueling} is another significant development. In situations with the exponentially vast environment and continuous action space, DDPG \cite{lillicrap2015continuous} was proposed which uses the actor-critic method. Other approaches are still the center research topic worldwide.

\newcolumntype{R}{>{\RaggedRight\let\newline\\\arraybackslash\hspace{15pt}}X}
% \FloatBarrier
\begin{table}[ht]
	\setlength\extrarowheight{5pt}
	\setlength{\tabcolsep}{30pt}
	\centering
	\caption{Abbreviation Table}
	\label{my-label1}
	\begin{tabularx}{21pc}{@{}p{80pt}p{85pt}@{}}
		\toprule
		\textbf{Abbreviation} & \textbf{Definition} \\  
		\midrule
		RL & Reinforcement Learning   \\ 
		DRL & Deep Reinforcement Learning \\
		IoT & Internet of Things \\
		CNN & Convolutional Neural Network \\
		MDP & Markov Decision Process \\
		CPS & Cyber Physical System  \\
		DoS & Denial of Service \\
		DDoS & Distributed Denial of Service \\
		DQN & Deep Q-Network \\ 
		ML & Machine Learning \\
		SINR & Signal-to-inference-plus-noise ratio \\
		SDN & Software Defined Network \\
		CRN & Cognitive Radio Network \\
		WACR & Wideband Autonomous CR \\
		VANET & Vehicular Ad-Hoc Network \\
		UAV & Unmanned Aerial Vehicle \\
		POMDP & Partially Observable MDP \\
		ICMP & Internet Control Message Protocol \\
				\bottomrule
	\end{tabularx}
\end{table}

\subsection{Comparison of RL with other Machine Learning}
Machine learning can be classified as Supervised, Unsupervised and Reinforcement Learning. In supervised learning, the ML model tries to predict the dependencies between training data and actual answer about a problem asked about that data \cite{8260800}. Basically, the input is given and we know what the model should predict in this kind of learning. It learns based on example. While reinforcement learning is about learning the environment without example. RL is more human like learning approach in which learning does not require large data. Here the agent do not know the target labels. Unsupervised Learning uses unlabeled data to understand the pattern. On the other hand, RL learn through interaction with environment without any prior data.

\subsection{Why Reinforcement Learning in IoT}
IoT connects millions of devices over the network. IoT devices are extremely dynamic and they make a complex network \cite{ling2015application}. Supervised and unsupervised learning technique have been used in security for intrusion detection \cite{garcia2009anomaly,dua2016data, buczak2015survey, berman2019survey,biswas2018intrusion,xin2018machine}, detection of malware \cite{milosevic2017machine, kp2018short,rege2018machine, berman2019survey}, CPS attack detection\cite{ding2018survey} \cite{wu2019detecting} and also in privacy maintenance task of IoT \cite{xiao2018iot}. However, these techniques can not perform dynamic responses for security in IoT environment \cite{nguyen2019deep}. For example for any new and constantly evolving cyber attacks, supervised and unsupervised learning method first need to get the dataset of those attacks and then only find a solution by learning the data. Reinforcement learning is applicable in IoT environment for many reasons. The real-time dynamic environment can be monitored efficiently in a favorable way. RL can continuously learn new information to accommodate to different advanced settings \cite{wang2016survey, wang2019survey}. Some IoT environments are so complex that it is difficult to model it. RL minimizes the effort associated with simulating and solving such complex environment. Consider a complex IoT scenario and we have to come up with a model that can solve a problem in that environment. For using supervised and unsupervised method, first simulation is to be performed to generate dataset and then only dataset is used to train the model. However, reinforcement learning algorithm performs trial and error in the environment and learns a model. This minimizes the complexity involved in simulating and solving a problem in a complex environment.  Data collection for some IoT environment is extremely difficult. In such a scenario, there are no datasets to train the model using other machine learning techniques. RL is the only machine learning technique that can learn without prior datasets.

\subsection{Reinforcement Learning for Securing IoT Against Adversarial Learning environment}
Reinforcement learning is regarded as one of the best solutions for securing IoT against adversarial learning environment that incorporates the environment's behavior into the learning process concurrently \cite{caminero2019adversarial}. This salient feature of Reinforcement Learning offers IoT security against adversarial learning environment where large number of diverse IoT devices produce huge amount of bursty data or continuous data stream.

\section{Threats and RL based solutions in IoT Security}
The rapid development of smart and mobile devices has made significant growth in IoT usage in many areas. Nowadays, IoT is incorporated in many domains. Industrial, power, agriculture, vehicles, battlefield, homes \cite{li2011smart} are the common application domain of IoT.
However, IoT is facing security problems with growth in its usage. Privacy and security maintenance is crucial for IoT systems. IoT uses advanced technologies like radio-frequency identifications (RFIDs), wireless sensor networks, Bluetooth, Zigbee, and cloud computing. Privacy protection and securing the system from cyber attacks like DoS attacks, jamming, eavesdropping, malware, and virus injection \cite{andrea2015internet} is a most and at the same time a very challenging task.  Privacy leakage is another challenge for security in IoT \cite{roman2013features}. For instance, devices that collect and report the actions of elderly people in a smart old care home must have to avoid private information leakage to prevent any harm to elderly people from attackers. IoT system is susceptible to attacks like network, software and physical attack. In this paper, we mainly look at the following IoT threats.

\subsection{Denial of Service Attack}

IoT systems face cyber-attack like Denial of Service (DoS) attack resulting in selective forwarding and eavesdropping \cite{alaba2017internet}. In IoT system attacks, DoS is the most common attack \cite{alanazi2015resilience}. These attacks on IoT networks place serious threats to human life and direct or indirect financial losses. A Denial-of-Service attack is a serious and most prevalent attack in which the attacker modifies the connection of network in such a way that it becomes unavailable to its expected users. DoS attack is achieved by flooding the communication network with unnecessary traffic. A denial of service attack disables the service in the victim's side by sending notably huge sizes of packets. The attack traffic can use the large portion of available bandwidth resulting in the services not reachable to legitimate users. Another critical challenge for IoT security in the current scenario is protecting the system against distributed denial of service(DDoS) attack. DDoS is typically a DoS attack but is of distributed nature. This results in a compromise of a tremendous number of IoT devices at a time. In 2016, a DDoS attack performed by Mirai botnet \cite{antonakakis2017understanding} had affected around 65,000 IoT devices just within the early 20 hours \cite{woolf}. DoS attack obstructs the usage for the genuine user resulting in the unavailability of network resources. DDoS is the same kind of attack but the only difference is it is drilled from distributed sources. IoT devices have limited power capacity to leverage mechanisms to detect these denial attacks. Network entrance for IoT can be the place to apply detection and protection mechanism from such attacks.
\\*
\subsubsection{ IoT layers}

\begin{figure}[ht!]
	\centering
	\includegraphics[width=35mm]{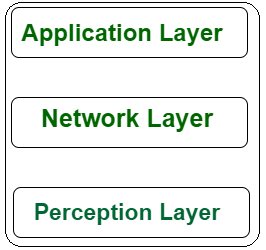}
	\caption{Three Layered IoT Architecture.}
	\label{IoTlayer}
\end{figure}

IoT architecture is mainly 3-layered \cite{inbook}. They are perception layer, network layer and application layer, as shown in Fig. \ref{IoTlayer}.

\begin{itemize}
	
	\item Perception Layer: The perception layer is all about sensing the physical characteristics of objects using sensors, actuators and other devices. The process of this perception is reliant on sensing technologies like RFID, GPS, 2-D barcode labels and readers \cite{wu2010research}. This layer also takes control of converting sensed information to digital signals. Chips to sense are to be designed and made as small as possible to implant it inside the tiny IoT devices. The main task of this layer is to gather the information by sensing the objects.
	\\*
	\item Network Layer: The network layer can be visualized as the neural network or brain of IoT. This layer is accountable for processing the information gathered from the Perception layer \cite{wang2016internet}. Also, it is responsible for transmitting the information to the application layer using wired/wireless networking technologies. Technologies like Wifi, Bluetooth, Zigbee, WirelessHART, Ethernet, 3G and so on are used to transmit the information. Because IoT sensors collect a massive amount of data, it is necessary to have a middleware that can handle this huge amount of data. For this, cloud computing is the main technology used in this layer.
	\\* 
	
	\item Application Layer: Application Layer is the topmost layer in IoT architecture which is the frontend of the IoT architecture. This layer realizes the application of the overall IoT system.  It supports by providing the demanded tools for developers to practicalize IoT vision. The application layer uses the data transmitted to them from previous IoT layers. Automatic sensing device management and node management are handled by this layer \cite{zhong2015study}. 
\end{itemize}

\subsubsection{DoS Attack in IoT layers}

\begin{itemize}
	\item DoS in Perception Layer: RFID is the main sensing technology used in the perception layer. Several attacks like Jamming\cite{finkenzeller2009known}, Kill Command Attack\cite{mitrokotsa2010classification}, De-synchronizing attack\cite{chien2007security} are common in this layer.

	\item DoS in Network Layer: Attackers perform flooding attacks like ICMP flood attack, Amplification based flooding, Reflection based flooding and many more\cite{sonar2014survey}. For instance, Wifi, which is the major technology in this layer suffers ICMP flooding attack.
	
	\item DoS in Application Layer:
	A common attack in the Application layer is Path based DoS attack\cite{deng2005defending}, Reprogramming attacks and so on.
\end{itemize}

\subsubsection{Reinforcement Learning against IoT DoS attack}

The authors in\cite{malialis2013multiagent} proposed an approach  to protect against DDoS attacks by using a Multiagent Router Throttling. They proposed a model where multiple reinforcement learning agents are involved. These agents are installed on routers. The agents learn to rate-limit or throttle traffic towards a victim server. It has been illustrated to work fine against DDoS attacks in small-scale network topologies. But this method suffered from scalability problems. To eliminate this issue, they proposed Coordinated Team Learning design on their multi-agent router throttling method \cite{malialis2015distributed}. This paper is centered on resolving the scalability issues as mentioned earlier. Here they have proposed an approach that combines mechanisms like hierarchical team-based communication, task decomposition, and team rewards to minimize the DDoS attack traffic. They referenced a network model as used by authors in \cite{yau2005defending} to develop emulator for throttling approaches. By using up to 100 reinforcement learning agents, the scalability of the proposed approach is evaluated. This method is applicable in highly scalable IoT environment. Simulation results showed that the adaptability of the proposed model is highly improved. Rl agents throttles the attacker traffic from flooding the server. Server is an important component in IoT mechanism. This approach minimizes  the DoS attacks in the server.

Software Defined Network (SDN) is a well known architecture for controlling large network space. SDN allows network administrators to have more control of the network and facilitate the efficient use of network resources \cite{hu2014survey}. SDN supports the separation of data plane and control plane in switches and routers \cite{fang2013loss}. The combination of IoT and SDN, commonly known as software defined internet of things, has a potential solution to managing IoT network traffic.The work in \cite{liu2018deep} tried  to mitigate DDoS attack using DDPG method which is more scalable than the work proposed in \cite{malialis2015distributed}. In this approach, the DRL agents are placed in the central Software Defined Network (SDN) instead of distributed router locations.  The DRL agent proposed here takes control of the traffic that reaches the server and prevents over flooding of traffic in the server. The mitigating agent is trained using DDPG algorithm and its state space are features of each port of switch and flow statistics. The authors have taken eight features in this paper. Action taken by an agent is throttling of traffic based on the maximum bandwidth allowed for a specific host. The DDoS attack mitigating agent gets a negative reward if it overloads the server with massive traffic. Also, it gets a reward based on the percentage of benign traffic and attack traffic reaching the server. The agent learns continuously and can take control of the traffic flowing to the server. Hence, it achieves the goal of mitigating DDoS attack on the server. The proposed agent can mitigate DDoS flooding attacks of different protocols such as TCP SYN, UDP and ICMP.

\subsection{Jamming Attack}
Jamming is an attack in which an attacker contaminates the original content of information by assigning interruption signals in the network or by barring the original content of the information \cite{8276173}. This results in the original content not reachable to the desired destination. Jamming is similar to a DoS attack. In wireless networks, the Jamming attack is achieved by decreasing the signal-to-noise ratio at the receiver side. This is achieved by passing interfering wireless signals to the network. The jamming attack can hinder the transmission of information between sender and receiver. Jammers use intentional radio interference to create disturbance in the network. This keeps the communicating medium busy not allowing the transmitter to transfer messages. The jamming attack can be proactive and reactive \cite{grover2014jamming}. In proactive jamming, jammers send the interference signal all the time without taking care of whether there is communication going on in the network. On the other hand, reactive jammers only attack when they sense communication in the network. Intelligent technologies like RL can be the potential research solution to jamming attacks in such IoT networks.
\\*

\subsubsection{Jamming Attack in IoT}

IoT is the large scale interconnected system that is vulnerable to numerous attacks due to its large attack surfaces. People are dependent on IoT devices more than ever and any attack on this system is serious to human life in some way.  The jamming attack is another serious attack in IoT that can severely disrupt the normal working of the IoT system. Jamming is one of the most dangerous attacks that can interfere in wireless communication channels in the network by injecting false packets and interrupting the radio communication frequencies. Considering this, the jamming attack is a major challenge and threat to IoT networks having nodes with confined energy and power \cite{weber2010towards}. Reactive jamming is a challenging attack faced by IoT networks compared to another jamming. Reactive jamming consumes the energy of low power devices unnecessarily. Thus, IoT devices being low power are mostly affected by this kind of attack.
There are many antijamming techniques proposed for general wireless network \cite{wu2011anti,el2014power,cagalj2006wormhole,wang2011anti}. The anti-jamming solution proposed for this traditional network is not applicable in the IoT network. The reasons are IoT network is highly dynamic, heterogeneous and more demanding. Also, IoT has limited memory, power, and transmission resources \cite{7841922}. More robust technologies like machine learning  can be an effective antijamming solutions in IoT environment.
\\*

\subsubsection{Reinforcement Learning against IoT Jamming attack}

IoT technology can perform well only if the communication of information is secure and efficient. So, the demand of wireless medium to support IoT functioning is high. It is challenging to properly assure the management and availability of spectrum resources. Unavailability of spectrum resources can impose a challenge to the sustaining of IoT technology. Cognitive Radio Network (CRN) in IoT somehow manages the spectrum utilization process. But the jamming attack is a serious security threat faced in CNR based IoT devices. Due to limited powered devices, wireless based IoT systems are more suffered by the jamming attack. Several anti-jamming algorithms has been proposed \cite{6523802, wang2011anti,8320276,heo2017dodge,kim2015cognitive,rawat2015securing,djuraev2017channel,becker2014dynamic}. In our paper, we briefly discuss the anti-jamming technique implemented using RL.
The work in \cite{8417695} proposed a deep reinforcement learning based power control scheme for IoT transmission against jamming. Convolution Neural Network is used as a deep learning algorithm.  The DQN-based power control scheme is implemented over the universal software radio peripherals (USRPs). Depending on the present IoT transmission status and strength of the jammer, the agents determine the transmit power unaware of the IoT topology. This approach showed enhanced signal-to-interference-plus-noise (SINR) of the IoT signals compared with anti-jamming using Q-learning. They have used DQN as an RL algorithm. Agent is the transmitter whose action is to choose and set the transmit power. On taking action, the SINR at a time slot is measured and calculated at the end.

The authors in \cite{7952524} have proposed a two dimensional anti-jamming communication using DRL. CRN is the network model used that has multiple Primary Users (PU) and jammers and a single Secondary User  (SU). In this scheme, SU, without interfering with PUs, utilizes both spread spectrum and user mobility to perceive jamming attacks. The authors proposed DQN based scheme to suggest the SU to take one of the two possible actions. First is to leave an area of heavy jamming and reconnect to another base station. Second, use one of the channel to send signals (frequency hopping) to beat the smart jammers. SU obtains an optimal anti-jamming communication policy by using DQN algorithm without having information about the jamming model and radio channel model. They used Convolution Neural Network to accelerate the learning rate. SINR and utility of the SU against cooperative jamming are improved compared to other learning approaches. SU is the agent that choose action based on the system state. State space includes the availability of the number of Primary Users and the discrete SINR value of the SU at that time slot. DQN approach followed in this work converged faster than Q-learning approach. The proposed DQN teaches the SU to choose optimal frequency hopping policy and hence mitigate a jamming attack. The CRN model has many application in IoT \cite{khan2017cognitive}. Thus the proposed RL based anti-jamming technique can be applicable in antijamming in CRN based IoT devices.

Alternatively, the authors in \cite{xiao2018anti} have proposed a model in which the receiver of SUs can decide to stay or leave the current location to combat jamming attacks. This mobility can cause some overhead, so it should find an optimal policy either to stay at the current location or move. Here, DQN based on CNN have been used by the receiver to choose the action that maximizes its utility. RL state space is the discrete measure of SINR of the signal sensed by the receiver at that time slot. Action by the receiver is whether to leave the location or stay there. They concluded that the proposed method achieved faster convergence and higher SINR as compared to Q-learning approach.

Both the works in \cite{7952524} and \cite{xiao2018anti} took account of the discrete SINR value as RL state. But in a scenario of infinitely large SINR, these approach is not suitable. Also, the SINR considered in these approaches may be noisy and false. To address this issue, a Recursive Convolutional Neural Network (RCNN) that handles infinite state problem was proposed by authors in \cite{8314744}. An optimal anti-jamming strategy was achieved by the proposed DRL model. The proposed algorithm improved the anti-jamming strategies against dynamic and intelligent jammers. Spectrum waterfall is defined as a state space of the RL environment. Spectrum waterfall utilizes the spectrum information with temporal features. It does not require jamming pattern information so it is applicable against smart jammers who continuously change their jamming pattern. The preprocessing layer in RCNN can remove excess noise from the environment and hence reduce complexity. It filters out the SINR with the help of a noise threshold. And recursive convolution layer handles the recursive input state. The simulation result validates the proposed algorithm by showing that the user can avoid jamming even if the jammers change jamming pattern intelligently. The proposed algorithm of DQL with RCNN shows faster convergence that Q-learning against fixed jamming attacks. The proposed method converged in the presence of dynamic jammers but Q-learning could not converge in this case. However, the work in \cite{li2019performance} provided a theoretical proof of a condition in which the method proposed by \cite{8314744} cannot converge. Here the authors raise a question against the previous DRL-based anti-jamming strategy. When the jammer is intelligent enough that can learn the communication pattern of the user and modify its jamming pattern accordingly, the previous model fails to converge. Here they design an RL agent against DRL anti-jamming. RL agent observes the frequency spectrum and based on that it chooses the frequency band to jam. They have opened a research challenge against intelligent jamming attacks.

Wideband autonomous cognitive radio (WACR) based antijamming using RL was proposed by authors in \cite{7636793}. WACR makes the use of its spectrum sensing ability to locate sweeping jammers. WACR not only senses the active signal but can also classify the signal properties which aids in finding such signals \cite{bkassiny2012wideband}. They define three steps wideband knowledge spectrum acquisition framework. They are wideband spectrum scanning, spectral activity detection, and signal classification. A reinforcement learning based decision policy is proposed in which a WACR learn an optimal policy to pick the sub-bands for sensing and transmission. The selection of sub-band is based on the desired contiguous length of idle bandwidth for a sub-band. For the sensed sub-band, Neyman-Pearson (NP) detector is used which allows the WACR to find the frequencies of all active signals in that sub-band. In this Q-learning based RL setting, the action of WACR is either to remain in a sub-band or to switch to other sub-band. WACR on taking each action updates its Q-table on the basis of reward it gets from the environment. Reward, in this case, is dependent on the amount of time WACR can avoid the jammer. Experimental results from the simulation showed that the Q-learning can learn the sweeping jammer pattern and can optimally switch the sub-bands to avoid jamming. A similar Q-learning approach is proposed for the WACR network in the work \cite{7925694}. The only difference is that the later one uses a multi-agent Rl approach. They considered multiple WACRs and proposed a similar Q-learning approach to achieve anti-jamming against sweeping jammers and interference from other WACRs. When multiple WACRs are operating in the same spectrum range and there is sweeper jamming, the proposed multi-agent approach avoids sweeping jammers and interference from other WACR. However, both \cite{7636793} and \cite{7925694} assumed fixed jammer. Both did not cover a scenario in which sweeping jammers can also be cognitive and smart enough to adjust its jamming strategy accordingly.

\subsection{Spoofing Attack}
A spoofing attack is a case in which a malicious node impersonates to be another person or device over the network. The main aim of this attack is to get trust from nodes and access the legitimate node to steal information or spread malware \cite{babu2010comprehensive}. Spoofing attackers trick the user or a node to believe that they are trustworthy and falsely access the information. A spoofing attack is of different types and we will discuss some of them in brief. IP spoofing is done by impersonating the IP address, sending information through that address and trick the receiver to believe that information \cite{4358709}. ARP spoofing is about sending the falsified ARP messages in the network \cite{whalen2001introduction}. The target of this attack is to falsify the victim node to send the information to a malicious node instead of sending it to a legitimate one. Other spoofing attacks like DNS spoofing, web spoofing, email spoofing, etc are common.

\subsubsection{Spoofing Attack in IoT}
IoT devices are interconnected and they share the information which is privacy critical. The taxonomy of security attacks in the work \cite{7804660} showed different cyber-attacks in IoT. Spoofing attack is a serious attack which may even lead to DDoS attack and Man-in-the-Middle attack. Let us look at a typical example of how spoofing attacks can disrupt IoT setting. Suppose an IoT scenario of connected multiple UAVs that are deployed in monitoring and controlling battlefield information. Spoofing attackers can be any unknown UAV that tries to join the network. On gaining trust from the network, the attacker can fake themselves to be genuine however it is malicious. The attacker UAV on joining the network can sense all the critical information of the battlefield. It can also pass false information in the network which will cause a serious disruption of the battlefield.

\subsubsection{Reinforcement Learning against IoT Spoofing Attack}
Reinforcement learning is like a game in which the agent plays with the opponent and learn the strategy followed by the opponent. In case of IoT communication, physical layer information like received signal strength, channel state information and channel impulse response can be useful in authenticating the transmitter \cite{liu2011robust}. Active authentication based on ambient radio signal is one way to authenticate the device and prevent spoofing attack. However, it is hard to obtain the dynamic time-variant channel mode in a real environment. Reinforcement learning was used to obtain this time-variant channel information in the work \cite{liu2017active}. Here the authors proposed an active authentication of mobile devices in the indoor environment using reinforcement learning. Here authors considered the PHY-layer information to detect spoofing attacks. The received signal strength at the receiver, which is trust authority, was considered to detect spoofing attack. The receiver formulates a hypothesis test to determine whether a packet is sent from the particular address or not. The receiver on getting a packet computes the test statistics of the hypothesis test. If it is below a threshold, the receiver accepts the packet as authentic otherwise detects the packet as a spoofed packet. The test threshold of hypothesis test in a dynamic environment is chosen by reinforcement learning. Q-learning was used to find the optimal threshold strategy without knowing the model of the arriving packet. State space in this environment is the false alarm rate and miss rate. Based on the observation of states, the receiver chooses a test threshold from L levels. State-action function in Q-learning is updated and utility calculated by the receiver is the reward function. Here the agent uses epsilon-greedy policy to get the optimal test threshold. The simulation result in experimenting with a legitimate user and three spoofers showed that the proposed Q-learning based test threshold strategy gave better utility. Also, the result showed that the proposed approach minimized the convergence of the false alarm rate and miss rate as contrasted to the fixed threshold approach. Compared to the fixed threshold approach, the proposed Q-learning-based threshold can efficiently detect the spoofing attacks in a dynamic environment.

The work in \cite{7398138} followed a similar approach as done by authors in \cite{liu2017active} to detect spoofing attack. Reinforcement learning was implemented to find the optimal test threshold. But here the authors have compared the performance of the RL agent on the following two algorithms. They have compared the performance of Q-learning and Dyna-Q algorithm. The simulation was implemented in an indoor environment in USRPs.  False alarm rate and miss rate are the states for the agent and utility at receiver is given as reward function. The simulation result showed that the error rate with Dyna-Q is lower than with Q-learning. The detection rate using both algorithms is better than the fixed threshold approach. Spoofing attack detection in an indoor environment is covered by previous papers. Authors in \cite{xiao2019learning} proposed a rogue edge detection scheme for VANETs (Vehicular Ad Hoc Network) observing the ambient radio signals. Similar to the previous approach, here also authors used Q-learning to allow mobile devices to reach optimal rogue edge attack detection policy without being aware of the dynamic VANET model.

Most of the security approaches are reactive i.e. they try to detect the security breach and then only recover from the attack. However, the work in \cite{bezzo2018predicting} tried to predict the intention of attack. They considered sensor spoofing attacks in one or more sensors of an autonomous vehicle with multiple sensors. The attackers try to take the vehicle in the undesired state by spoofing and hiding inside the sensors. Here authors came up with a Reachability-based approach and Inverse RL to predict the intention of the attacker and detect the compromised sensors. First reachability analysis was used, as done by authors in \cite{8046382}, to find the set of possible states the vehicle can reach on a certain time slot. Inverse RL was used to infer the maximum reward function expected by the attacker. The approach used was to find the group of sensors that deviate the vehicle towards the undesired state. Bayesian Inverse RL (BIRL) can learn the reward function in Markov Decision Process (MDP) if they are given the behavior and dynamics of the system \cite{ramachandran2007bayesian}. Given the set of observations, they calculate the posterior probability of all reachable goals. If any of the sensors return a state value such that the variance of the posterior probability is within the user selected threshold, the recovery procedure is initiated. The simulation result showed that when the variance of the posterior probability of goal of attacker reached below the threshold value, the spoofed sensor was detected and omitted from the state estimation and a recovery process is initiated. Similar to this work, BIRL was used in the work \cite{elnaggar2018irl} to detect the spoofed sensor in an autonomous vehicle environment. In this work, authors have used active exploration policy in which the vehicle explores the environment to reach sensitive states. Active exploration prevented the vehicle to reach states very near to the undesirable state.

\section{Reinforcement Learning in Cyber Physical Systems}

\subsection{Security in Smart Grid}
Smart Grid is an intelligent system to generate and distribute energy in a distributed manner. It is the combination of the traditional power grid and information systems that allows efficient energy generation and consumption. Digital processing in the traditional power grid leads to Smart Grid which gives the capability to control, communicate and monitoring of available energy sources. However, smart grids being online and connected is vulnerable to various cyber attacks. The integration of cyber component exposes it to critical cyber-attacks and unauthorized penetration. Cyber Physical Attack also called a blended attack imposes a threat to both the cyber and physical systems of the grid thus causing negative consequences than by the individual attack (cyber attack or physical attack) \cite{pillitteri2014guidelines}. Attacks like information tampering and eavesdropping throw a big threat to the security of Smart Grid \cite{wang2013survey}. Researchers around the world are concerned about the security in this area of CPS and have proposed several security approaches. Here, we will talk about security approaches taken to protect Smart Grid using RL.

A sequential attack on the network topology of a smart grid is a serious attack in which the attacker can determine the number and time to attack the component to cause maximum damage. A sequential attack imposes more damage than by a simultaneous attack when attacked on the same victim links \cite{zhu2014sequential}. The authors in \cite{yan2016q} proposed a Q-learning based vulnerability analysis of smart grid under sequential attack admitting the physical system behaviors. Here the authors defined sequential attack as a sequence of coordinated interdiction such that it changes an in-service line into out-of-service. By manipulating the control commands or false line status data such an attack can be performed resulting in cascading blackout. The authors proposed a Q-learning based vulnerability analysis in a smart grid under sequential attack. In this RL environment, the agent is the attacker who tries to identify the more vulnerable point in the grid under a sequential attack. State space is either in-service line or out-of-service line at a time. Action taken by the attacker is to maliciously turn the in-service line to out of service. The goal of the attacker using Q-learning is to mind the optimal policy to fail the system with the least number of lines attacked. If the attacker can reach blackout by turning of lines equal to or more than a threshold line value and with less action than the threshold value, it will get a positive reward. It gets a negative reward on taking more action than the threshold. Otherwise, the reward given is zero. The experimental results of this approach successfully identified the critical sequential topology attacks.
The blackout sizes in the proposed method showed that with the increase in load, sequential attacks caused more line outages and attack intentions were accomplished quicker. The proposed Q-learning approach tried to learn and find out vulnerable sequences that directed to severe blackouts in the system. Using this vulnerability analysis of the sequential attack, the defender side can follow the security measures to better the situational awareness cyber-security approaches.

False Data Injection (FDI) is proven to be a challenging attack to the smart grid in which the attacker injects malicious data to the Supervisory Control and Data Acquisition (SCADA) system resulting in cascading failure of a smart power system. In the work \cite{chen2018evaluation}, an intelligent FDI attack on a smart grid with automatic voltage control was studied. The authors considered a smart attacker that uses Q-learning approach to find the optimal attack strategy stealthily to manipulate the control system in a compromised substation. In the given paper, the state space is the voltage angle, the amplitude of the buses, the active and reactive power of the generator and, the active and reactive power of the load. An attacker can perform FDI only based on local observation so authors considered the attack as Partially Observable MDP (POMDP). Attacker action is to compromise many measurements of the attacked substation. The reward function is defined such that the attacker's action makes the bus voltage in compromised substation lower than the desired operational voltage. By giving rewards for no action, the attacker stops injecting FDI and avoids giving a fixed action pattern. Using Q-learning method with the nearest memory sequence results showed that the proposed FDI, with little knowledge of the complete power system, could generate voltage breakdown events. Online learning helps the attacker to choose probable attack times automatically to make the attack silent. The test result shown in the results section validated the bad data detection and correction method presented against the proposed FDI attack.

Some advantages and disadvantages of the some approaches for securing IoT with reinforcement learning is tabulated in Table \ref{my-label2}.

\newcolumntype{R}{>{\RaggedRight\let\newline\\\arraybackslash\hspace{0pt}}X}
\begin{table*}[ht!]
	\setlength\extrarowheight{5pt}
	\centering
	\caption{Advantages and Disadvantages of the some of the research work associated with securing IoT with reinforcement learning}
	\label{my-label2}
	\begin{tabularx}{\textwidth}{|R|R|R|R}
		\toprule
		\textbf{Approach} & \textbf{Goal} & \textbf{Specialities(+) and Limitations(-)} \\
		\midrule
		Multiagent Router Throttling \cite{malialis2013multiagent} & Multiple Agent Learn to throttle the traffic to victim server & \makecell[tl]{+ Solves Stability issue \\  -Not Scalable} \\
		\hline
		
		Coordinated Team Learning in Multiagent Router Throttling \cite{malialis2015distributed} & Hierarchical team based communication to throttle excessive traffic reaching server & \makecell[tl]{+ Scalability is achieved \\  + Improved Adaptability \\ - Consideration of less statistical feature \\ -Lower Data Efficiency} \\
		
		\hline
		
		Smart mitigation Agent in Software Defined Network \cite{liu2018deep} & Mitigation Agent throttles the traffic by evaluating the controller of SDN & \makecell[tl]{+Highly Scalable \\ +Improved Data Efficiency \\  +Reduces overhead on SDN switches}  \\
		
		\hline
		
		Power Control for IoT against Jamming \cite{8417695} & IoT device decides transmit power in a way to improve SINR and utility in the presence of jammers & \makecell[tl]{+More realistic approach \\ (experimented under hardware constraint) \\  +Improved Communication Efficiency \\  - Cost overhead} \\
		
		\hline
		
		Two-dimensional anti-jamming communication \cite{7952524} & Avoid jamming attack smartly without interfering primary user & \makecell[tl]{+Faster Convergence \\ - Cannot handle smart jammers} \\
		
		\hline
		
		Antijamming in underwater acoustic network \cite{xiao2018anti} & To control transmit power against jamming in acoustic network for underwater robots and vehicles. & \makecell[tl]{+ Higher learning speed \\ - Not scalable \\ - Cannot handle smart jammer} \\
		
		\hline
		
		Antijamming communication using Spectrum Waterfall \cite{8314744} & To achieve antijamming in dynamic environement in the presence of smart jammer & \makecell[tl]{ + Less information loss \\ + Reduced complexity \\ - Cannot converge in the presence of \\ \ \ RL-based jammer} \\ 
		
		\hline
		
		Antijamming with Wideband Autonomous Cognitive Radio \cite{7636793} & Optimal sub-band selection against jammers & \makecell[tl]{+Reduced Complexity \\ - Not practical} \\
		
		\hline
		
		Active Authentication of mobile devices \cite{liu2017active} & Autheticate mobile devices against spoofing attack & \makecell[tl]{+ Privacy Protection \\ + Reduced overhead cost \\ - Not Scalable}\\ 
		\hline
		
		Physical Layer Rogue edge detection in VANET \cite{xiao2019learning}  & To find rogue edge node based on physical properties of ambient radio signals & \makecell[tl] {+Handle dynamic environment} \\
		\hline
		
		Predicting malicious intention under cyber attack \cite{bezzo2018predicting} & To predict the goal of sensor spoofing attack and determine the compromised sensor & \makecell[tl]{ +More realistic approach \\ -Higher Computation Complexity \\ -Slower convergence speed} \\  
		\hline
		RL approach for attack intention prediction \cite{elnaggar2018irl} & To predict the intention of attacker and detect the set of compromised sensor & \makecell[tl]{+Faster convergence speed \\ - Complex computation } \\
				
		\bottomrule
			
	\end{tabularx}
\end{table*}

The work in \cite{8514804} tried to design a defender system against cyber attack in a smart grid using reinforcement learning. The authors proposed a model free RL algorithm that can defend cyber attacks on the fly without knowing any attack model. A defender is proposed that can detect the low magnitude attack which will be the worst-case scenario for the defender. This makes the defender sensitive to even a very slight deviation of measurement from a normal measurement. The proposed defender system also limits the action space of the attacker. An attacker can only make a lower magnitude attack to be not detected. Such a lower magnitude attack, however, can not make damage to the system. The agent does not know the attacker attack time, so they considered two state i.e. preattack and postattack state. State space is the status of transmission lines in the power system. After observing the measurement, agent (defender) can take two actions. Either they can stop and declare an attack or they can continue to obtain more measurements. The goal of the agent to lessen the detection delays and false alarm rates. Here the reward is the cost associated with the detection delay compared with the false alarm rate. If the agent in preattack state takes action to stop, it gets a unitary reward. While if in postattack state it takes continue action, a cost is given as a reward which is due to the detection delay. SARSA algorithm was used to train the agent and update the Q- table. SARSA is a model free RL algorithm that is shown to have better performance in POMDP environment \cite{loch1998using}. Using this learned Q-table, the agent performs online attack detection by choosing the action that leads to the minimum expected future cost. The agent continues to take action until it takes stop action on which it declares that there is attack in the system. They have shown the simulation result of the proposed RL based defender in the presence of different kinds of attacks and compared with Euclidean detector and Cos-Sim detector. The result showed that RL based detector detected the attack with very low detection delay as compared to other approaches. However, here they consider single agent defender which can be extended to multi agent and they have not considered a smart attacker.

Ni and Paul \cite{8603817}  proposed a dynamic game between the attacker and the defender to find the optimal attack strategies using reinforcement learning. The attacker learns the attack sequence to be applied in the transmission lines. On the other hand, the defender learns to protect the lines selected. An attacker takes generation loss and line outages as the reward and based on which it plans the next action. Here the attacker finds the critical transmission lines in the smart grid based on the action taken by the defender. This learned attack sequence is used by the defender so that it minimizes the action set for the attacker.
In this multistage game, they first assumed defender to be passive and attacker to be smart learner. Defender policy was predefined so using that policy information, the attacker performs trial and error action using Q-learning and conducts an attack on the transmission line. Calculation of generation loss and cascade are done and after getting a reward, the Q-table is updated. Later at the end of this multistage game, the defender aligns its action based on the observation and using the sequence learned by the attacker. Here the game was proposed as a zero-sum game in which the reward given to attacker and to defender are opposite to each other. The experimental result showed that total line outages caused by the multistage attack are more consequential than a single stage attack. Also, the result here showed the decreased number of successful attacks and average generation loss on adjusting with the strategy of the defender. The shown case studies in the paper imply that learned information of the attacker can ultimately assist the defenders to plan for better defense policies.

\subsection{Security in Smart Transportation System}

Smart Transportation System (STS) is a CPS system that consists of sensors technology, control, and communication in vehicles and any other transportation infrastructure. The goal is to provide real-time road and other vehicle information for users to improve safety and comfort in transportation. It achieves the smartness in transportation by establishing the connection between vehicle to vehicle (V2V), vehicle to other infrastructures (V2I), vehicles to pedestrian and so on \cite{karagiannis2011vehicular}. However, security challenges in STS are posing a threat to this system. It should properly handle issues like privacy protection, authorization, data integrity, data storage, and management \cite{8855737}. Security is always perceived as one of the most important considerations in realizing STS usecases \cite{javed2016security}. Cybersecurity researchers around the world have proposed several methods to secure this transportation. Machine Learning is an emerging technology that have added more smartness to the STS system and also it has been used in securing the system intelligently. VANET and FANET (Flight Ad-Hoc Network) can both be considered as STS.

% \begin{figure}[ht!]
%	\centering
%	\includegraphics[width=90mm]{figure/SMARTCITY.png}
%	\caption{Smart Transportation for Smart City Conceptual Model %\cite{javed2016security}\label{smart1}}
%\end{figure}

%\newcolumntype{R}{>{\RaggedRight\let\newline\\\arraybackslash\hspace{0pt}}X}
\begin{table*}[t!]
	\setlength\extrarowheight{5pt}
	\centering
	\caption{Reinforcement learning parameters followed by some of the research approaches.}
	\label{my-label3}
	\begin{tabularx}{\textwidth}{|R|R|R|R|R|R|}
		\toprule
		\textbf{References} & \textbf{Algorithm} & \textbf{Agent} & \textbf{State-Space} & \textbf{Action-Space} &	\textbf{Reward} \\
		\midrule
		\cite{malialis2013multiagent} & SARSA & Router & Traffic flowing towards server & Probabilistic throttle of traffic coming from host & Negative reward if server is overloaded otherwise reward dependent on the rate of legitimate traffic \\
		\hline
		
		\cite{liu2018deep} & DDPG & Mitigating Router & Flow statistics and features of each port & Generate a vector representing maximum bandwidth of specific host & Negative reward if load on server exceeds an upper bound otherwise defined by reward function   \\
		\hline
		\cite{8417695} & DQN with CNN  & IoT devices & SINR and utility value & Decide transmit power at a time slot & Positive reward if SINR and utility is improved \\
		\hline
		\cite{7952524} &DQN with CNN  & Mobile device & Presence of PUs and SINR of the signal at previous timeslot & Decide to leave or stay at an area and choose channel & SINR and utility \\
		\hline
		\cite{xiao2018anti} & DQN with CNN & Sensor & SINR and RSSI at previous timeslot & Choose trasmit power and decide to stay or move to another area & Utility of the signal \\
		\hline
		\cite{8314744} & DRL with RCNN & Sensor &  Raw frequency spectrum information & Choose signal frequency & Defined by function based on utility and cost of frequency switching \\
		\hline
		\cite{7636793} &Q- learning & Radio & Required bandwidth length & Select a new sub-band & Time taken by jammer to interfere the transmission after switching to a sub band \\
		\hline
		\cite{liu2017active} & Q-Learning & Radio Device & False rate and miss rate of authenticated packets & Select test threshold value & Utility \\
		
		\bottomrule
	\end{tabularx}
\end{table*}

Next we discuss the application of RL for security in STS. 
The work in \cite{8231220} presented deep reinforcement learning  approach for Unmanned Aerial Vehicle (UAV) against smart attacks with no information on the attack model and accuracy of the system to detect the attack. DQN was used to find the optimal power allocation strategy against a smart attacker. Authors first formulated a prospect theory based smart attack game to find the attack on UAV transmission by a subjective attacker. Then DQN is proposed to find the optimal power allocation strategy in multiple frequency channels. They compared the convergence rate achieved by using Q-learning, DQN and WoLF-PHC (Win or Learn Fast- Policy Hill Climbing) for power allocation against the attacker. UAV sends a signal in each time slot with certain power using DQN approach and observing the state of the network. Observation is the SINR value and utility of the received signal. The results shown in the paper depicts that DQN based power allocation is applicable for UAV having enough resources. On the other hand, WoLF-PHC based strategy can choose a transmission strategy with a lower computational cost. Here UAV can address the Q-learning based smart attack by learning the optimal transmission strategy.
VANETs in a large network topology bring high mobility in the onboard units (OBUs). Due to this large scale and dynamic nature, an antijamming strategy like frequency hopping is not efficient. The work in \cite{8246580} presented a UAV based antijamming approach in VANET using reinforcement learning. This is a follow up research for the proposed UAV relay strategy in the work \cite{lu2017anti} considering more practical aspects. Here authors proposed a relay game in which UAV learns whether or not to relay the OBUs data to another roadside unit (RSU) and smart jammer decides its jamming power. The authors presented the Nash equilibrium (NE) to show the dependence between the transmission cost and channel model with the UAV relay strategy. Here hotbooting-PHC based strategy was presented for faster UAV relay decisions. Hot booting uses the experimental data generated in advance to update the Q-table. This initializes the Q-value and probability of action-state and hence learning speed is significantly higher. Here UAV decides relay action based on the observed Bit Error Rate (BER) of the data send by UAV and the channel quality. The experimental result showed the decreased BER of OBU data and increased utility of VANET using the proposed algorithm than by using the Q-learning approach. Thus if the serving RSU for an OBU in VANET is severely jammed, the proposed UAV based relay strategy can transfer that information to another RSU and prevents the VANET from potential jamming affects.

Connected and autonomous vehicles (CAVs) and UAVs in IoT can be misused by attackers which directly impose a threat to the STS. The authors in \cite{8761101} proposed an anti-jamming V2V communication in an integrated UAV-CAV network with hybrid attackers. They assumed a malicious CAV that can perform smart jamming and a malicious UAV without smartness. Inspired by the predictive-adaptation feature of the human brain, they proposed a research tool called CDS to lead the idea of an anti-jamming technique. The process of channel selection is based on the risk level evaluation by task-switch control and following the process of power control completion. Reinforcement Learning is used for power control and channel selection. Here, the channel selection task is viewed as multi-armed bandit (MAB) problem \cite{weng2018bandit} and the upper confidence bound (UCB1) algorithm (index based policy) \cite{auer2002finite} is used as its solution. Experimental results showed a better transmission power and channel allocation strategy against hybrid attackers by the proposed method.
Typical parameters for reinforcement learning for different research works are listed in Table \ref{my-label3}.

\section{ Research Trends and Open Research Challenges}
IoT is a highly dynamic environment generating a massive amount of transactions. Connected devices in the network can be millions in numbers making the security approaches more challenging.  Reinforcement Learning is proven to be applicable in securing the IoT technology from various attacks. We presented research works done to secure the IoT system in our paper. It has been proven a powerful technology in IoT security. However, there are some challenges to be considered. Some of the challenges and open issues are discussed below.
\subsection{Discretizing of action-state space and minimizing curse of dimensionality}

Most of the work done addresses finite action and state space i.e. a discrete set of action spaces or finite MDP problems. However, in a real IoT environment, the RL algorithm should take care of continuous action and state spaces. Several works have considered discretizing the action-state spaces. But discretizing is a very expensive learning process and not suitable in extremely non-linear problems like in IoT environment.  The actor-critic approach has been proven to be an effective solution to address continuous action space \cite{lazaric2008reinforcement}. Application of actor-critic algorithms in solving IoT security issues can be a further research approach in this field. Also, there is another approach called hierarchical deep reinforcement learning which decomposes the problem states into smaller parts \cite{kulkarni2016hierarchical}. This can reduce the curse of dimensionality as faced by traditional RL approaches. This approach minimizes scaling problem by sub tasking any task and hence minimizing action and state space at a time. In such a dynamic and huge IoT environment, a hierarchical RL approach can be applied for better and quick optimization. 

\subsection{Learn with partially observable environment}
Reinforcement learning is an optimization problem that considers MDP environment. However, in IoT scenario, most of the environment are only partially observable. The reinforcement learning agent in this IoT scenario, can not have the complete perception of the environment. The reasons are because sensors are of limited sensing capacity and there is transmission loss due to limited transmission capacity in IoT. DRL approaches have been used in POMDP environment. But it is only applicable is small scale IoT environment. The potential solution to this problem could be the integration of recurrent neural network and RL to find the policies in POMDP environment. 

\subsection{Joint reward from multiple agents}
We have discussed the research work that considered multiple RL agent located at different devices or sensors in a distributed manner.  Each agent can have specific task or similar task to perform. In most of the works, multiple agents are thought to perform similar task. The design of multi agent RL system with different agents performing different task is to be studied. The challenge could be the collaborative method of considering the rewards from all the agents. This makes the application of multiple agents a complex problem. IoT environment is highly dynamic and the proper control between different agents is a demanding task to be worked on.

\subsection{Robustness against Adversarial RL}
One of the challenge and active research area to apply reinforcement learning in IoT is the consideration of adversarial environment. Very few works in literature have looked into the problem of applying RL against adversarial conditions. The environment can be adversary which continuously tries to win over the agent trying to learn the environment. In an multi agent RL problem, one of the agent can be an adversary. Therefore there should be a way to ensure that the learned policy is robust against any uncertain changes in the environment. Methods to make the agents trained by reinforcement algorithm robust against any adversarial attacks is an open research challenge.

\section{Conclusion}
In this paper, we have presented a comprehensive survey on the application of Reinforcement learning for IoT security. First, we have given a brief introduction about Reinforcement learning and background information about several attacks in IoT. Following that, we have presented a survey on various reinforcement learning techniques proposed against IoT attacks such as jamming attack, spoofing attack and denial of service attack. Furthermore, we have presented the Reinforcement learning for securing CPS systems (i.e., IoT with feedback and control) such as smart grid and smart transportation system. Moreover, we have presented some open research challenges and some research direction for IoT security using reinforcement learning.

%\bibliographystyle{ieeetr}
%\bibliography{references_CPS}

\begin{thebibliography}{}

\end{thebibliography}


\begin{thebibliography}{100}
	
	\bibitem{atzori2010internet}
	L.~Atzori, A.~Iera, and G.~Morabito, ``The internet of things: A survey,'' {\em
		Computer networks}, vol.~54, no.~15, pp.~2787--2805, 2010.
	
	\bibitem{felix2020sur}
	F.~Olowononi, D.~B. Rawat, and C.~Liu, ``{Resilient Machine Learning for
		Networked Cyber Physical Systems: A Survey for Machine Learning Security to
		Securing Machine Learning for CPS},'' {\em IEEE Communications Surveys and
		Tutorials}, 2020.
	\newblock Early Access, DoI: https://doi.org/10.1109/COMST.2020.3036778.
	
	\bibitem{rawat2018performance}
	D.~B. Rawat, R.~Alsabet, C.~Bajracharya, and M.~Song, ``{On the performance of
		cognitive internet-of-vehicles with unlicensed user-mobility and licensed
		user-activity},'' {\em Computer Networks}, vol.~137, pp.~98--106, 2018.
	
	\bibitem{rawat2017vehicular}
	D.~B. Rawat and C.~Bajracharya, ``{Vehicular Cyber Physical Systems: Adaptive
		Connectivity and Security},'' tech. rep., Springer, 2017.
	
	\bibitem{mansfield2018open}
	S.~Mansfield-Devine, ``Open source and the internet of things,'' {\em Network
		Security}, vol.~2018, no.~2, pp.~14--19, 2018.
	
	\bibitem{min2018learning}
	M.~Min, X.~Wan, L.~Xiao, Y.~Chen, M.~Xia, D.~Wu, and H.~Dai, ``{Learning-based
		privacy-aware offloading for healthcare IoT with energy harvesting},'' {\em
		IEEE Internet of Things Journal}, vol.~6, no.~3, pp.~4307--4316, 2018.
	
	\bibitem{sutton2018reinforcement}
	R.~S. Sutton and A.~G. Barto, {\em Reinforcement learning: An introduction}.
	\newblock MIT press, 2018.
	
	\bibitem{5227780}
	F.~L. {Lewis} and D.~{Vrabie}, ``Reinforcement learning and adaptive dynamic
	programming for feedback control,'' {\em IEEE Circuits and Systems Magazine},
	vol.~9, no.~3, pp.~32--50, 2009.
	
	\bibitem{schultz1997neural}
	W.~Schultz, P.~Dayan, and P.~R. Montague, ``A neural substrate of prediction
	and reward,'' {\em Science}, vol.~275, no.~5306, pp.~1593--1599, 1997.
	
	\bibitem{barto2003recent}
	A.~G. Barto and S.~Mahadevan, ``Recent advances in hierarchical reinforcement
	learning,'' {\em Discrete event dynamic systems}, vol.~13, no.~1-2,
	pp.~41--77, 2003.
	
	\bibitem{hinton2006reducing}
	G.~E. Hinton and R.~R. Salakhutdinov, ``Reducing the dimensionality of data
	with neural networks,'' {\em science}, vol.~313, no.~5786, pp.~504--507,
	2006.
	
	\bibitem{mnih2016asynchronous}
	V.~Mnih, A.~P. Badia, M.~Mirza, A.~Graves, T.~Lillicrap, T.~Harley, D.~Silver,
	and K.~Kavukcuoglu, ``Asynchronous methods for deep reinforcement learning,''
	in {\em International conference on machine learning}, pp.~1928--1937, 2016.
	
	\bibitem{dreyfus2002richard}
	S.~Dreyfus, ``Richard bellman on the birth of dynamic programming,'' {\em
		Operations Research}, vol.~50, no.~1, pp.~48--51, 2002.
	
	\bibitem{franccois2018introduction}
	V.~Fran{\c{c}}ois-Lavet, P.~Henderson, R.~Islam, M.~G. Bellemare, J.~Pineau,
	{\em et~al.}, ``An introduction to deep reinforcement learning,'' {\em
		Foundations and Trends{\textregistered} in Machine Learning}, vol.~11,
	no.~3-4, pp.~219--354, 2018.
	
	\bibitem{7983780}
	S.~P.~K. {Spielberg}, R.~B. {Gopaluni}, and P.~D. {Loewen}, ``Deep
	reinforcement learning approaches for process control,'' in {\em 2017 6th
		International Symposium on Advanced Control of Industrial Processes
		(AdCONIP)}, pp.~201--206, 2017.
	
	\bibitem{mao2016resource}
	H.~Mao, M.~Alizadeh, I.~Menache, and S.~Kandula, ``Resource management with
	deep reinforcement learning,'' in {\em Proceedings of the 15th ACM Workshop
		on Hot Topics in Networks}, pp.~50--56, 2016.
	
	\bibitem{zhang2017intelligent}
	Y.~Zhang, J.~Yao, and H.~Guan, ``Intelligent cloud resource management with
	deep reinforcement learning,'' {\em IEEE Cloud Computing}, vol.~4, no.~6,
	pp.~60--69, 2017.
	
	\bibitem{vecerik2017leveraging}
	M.~Vecerik, T.~Hester, J.~Scholz, F.~Wang, O.~Pietquin, B.~Piot, N.~Heess,
	T.~Roth{\"o}rl, T.~Lampe, and M.~Riedmiller, ``Leveraging demonstrations for
	deep reinforcement learning on robotics problems with sparse rewards,'' {\em
		arXiv preprint arXiv:1707.08817}, 2017.
	
	\bibitem{gu2017deep}
	S.~Gu, E.~Holly, T.~Lillicrap, and S.~Levine, ``Deep reinforcement learning for
	robotic manipulation with asynchronous off-policy updates,'' in {\em 2017
		IEEE international conference on robotics and automation (ICRA)},
	pp.~3389--3396, IEEE, 2017.
	
	\bibitem{mnih2015human}
	V.~Mnih, K.~Kavukcuoglu, D.~Silver, A.~A. Rusu, J.~Veness, M.~G. Bellemare,
	A.~Graves, M.~Riedmiller, A.~K. Fidjeland, G.~Ostrovski, {\em et~al.},
	``Human-level control through deep reinforcement learning,'' {\em Nature},
	vol.~518, no.~7540, pp.~529--533, 2015.
	
	\bibitem{silver2016mastering}
	D.~Silver, A.~Huang, C.~J. Maddison, A.~Guez, L.~Sifre, G.~Van Den~Driessche,
	J.~Schrittwieser, I.~Antonoglou, V.~Panneershelvam, M.~Lanctot, {\em et~al.},
	``Mastering the game of go with deep neural networks and tree search,'' {\em
		nature}, vol.~529, no.~7587, p.~484, 2016.
	
	\bibitem{lillicrap2015continuous}
	T.~P. Lillicrap, J.~J. Hunt, A.~Pritzel, N.~Heess, T.~Erez, Y.~Tassa,
	D.~Silver, and D.~Wierstra, ``Continuous control with deep reinforcement
	learning,'' {\em arXiv preprint arXiv:1509.02971}, 2015.
	
	\bibitem{van2016deep}
	H.~Van~Hasselt, A.~Guez, and D.~Silver, ``Deep reinforcement learning with
	double q-learning,'' in {\em Thirtieth AAAI conference on artificial
		intelligence}, 2016.
	
	\bibitem{wang2015dueling}
	Z.~Wang, T.~Schaul, M.~Hessel, H.~Van~Hasselt, M.~Lanctot, and N.~De~Freitas,
	``Dueling network architectures for deep reinforcement learning,'' {\em arXiv
		preprint arXiv:1511.06581}, 2015.
	
	\bibitem{8260800}
	I.~{Kachalsky}, I.~{Zakirzyanov}, and V.~{Ulyantsev}, ``Applying reinforcement
	learning and supervised learning techniques to play hearthstone,'' in {\em
		2017 16th IEEE International Conference on Machine Learning and Applications
		(ICMLA)}, pp.~1145--1148, 2017.
	
	\bibitem{ling2015application}
	M.~H. Ling, K.-L.~A. Yau, J.~Qadir, G.~S. Poh, and Q.~Ni, ``Application of
	reinforcement learning for security enhancement in cognitive radio
	networks,'' {\em Applied Soft Computing}, vol.~37, pp.~809--829, 2015.
	
	\bibitem{garcia2009anomaly}
	P.~Garcia-Teodoro, J.~Diaz-Verdejo, G.~Maci{\'a}-Fern{\'a}ndez, and
	E.~V{\'a}zquez, ``Anomaly-based network intrusion detection: Techniques,
	systems and challenges,'' {\em computers \& security}, vol.~28, no.~1-2,
	pp.~18--28, 2009.
	
	\bibitem{dua2016data}
	S.~Dua and X.~Du, {\em Data mining and machine learning in cybersecurity}.
	\newblock CRC press, 2016.
	
	\bibitem{buczak2015survey}
	A.~L. Buczak and E.~Guven, ``A survey of data mining and machine learning
	methods for cyber security intrusion detection,'' {\em IEEE Communications
		surveys \& tutorials}, vol.~18, no.~2, pp.~1153--1176, 2015.
	
	\bibitem{berman2019survey}
	D.~S. Berman, A.~L. Buczak, J.~S. Chavis, and C.~L. Corbett, ``A survey of deep
	learning methods for cyber security,'' {\em Information}, vol.~10, no.~4,
	p.~122, 2019.
	
	\bibitem{biswas2018intrusion}
	S.~K. Biswas, ``Intrusion detection using machine learning: A comparison
	study,'' {\em International Journal of Pure and Applied Mathematics},
	vol.~118, no.~19, pp.~101--114, 2018.
	
	\bibitem{xin2018machine}
	Y.~Xin, L.~Kong, Z.~Liu, Y.~Chen, Y.~Li, H.~Zhu, M.~Gao, H.~Hou, and C.~Wang,
	``Machine learning and deep learning methods for cybersecurity,'' {\em IEEE
		Access}, vol.~6, pp.~35365--35381, 2018.
	
	\bibitem{milosevic2017machine}
	N.~Milosevic, A.~Dehghantanha, and K.-K.~R. Choo, ``Machine learning aided
	android malware classification,'' {\em Computers \& Electrical Engineering},
	vol.~61, pp.~266--274, 2017.
	
	\bibitem{kp2018short}
	S.~KP {\em et~al.}, ``A short review on applications of deep learning for cyber
	security,'' {\em arXiv preprint arXiv:1812.06292}, 2018.
	
	\bibitem{rege2018machine}
	M.~Rege and R.~B.~K. Mbah, ``Machine learning for cyber defense and attack,''
	{\em DATA ANALYTICS 2018}, p.~83, 2018.
	
	\bibitem{ding2018survey}
	D.~Ding, Q.-L. Han, Y.~Xiang, X.~Ge, and X.-M. Zhang, ``A survey on security
	control and attack detection for industrial cyber-physical systems,'' {\em
		Neurocomputing}, vol.~275, pp.~1674--1683, 2018.
	
	\bibitem{wu2019detecting}
	M.~Wu, Z.~Song, and Y.~B. Moon, ``Detecting cyber-physical attacks in
	cybermanufacturing systems with machine learning methods,'' {\em Journal of
		intelligent manufacturing}, vol.~30, no.~3, pp.~1111--1123, 2019.
	
	\bibitem{xiao2018iot}
	L.~Xiao, X.~Wan, X.~Lu, Y.~Zhang, and D.~Wu, ``Iot security techniques based on
	machine learning: How do iot devices use ai to enhance security?,'' {\em IEEE
		Signal Processing Magazine}, vol.~35, no.~5, pp.~41--49, 2018.
	
	\bibitem{nguyen2019deep}
	T.~T. Nguyen and V.~J. Reddi, ``Deep reinforcement learning for cyber
	security,'' {\em arXiv preprint arXiv:1906.05799}, 2019.
	
	\bibitem{wang2016survey}
	W.~Wang, A.~Kwasinski, D.~Niyato, and Z.~Han, ``A survey on applications of
	model-free strategy learning in cognitive wireless networks,'' {\em IEEE
		Communications Surveys \& Tutorials}, vol.~18, no.~3, pp.~1717--1757, 2016.
	
	\bibitem{wang2019survey}
	Y.~Wang, Z.~Ye, P.~Wan, and J.~Zhao, ``A survey of dynamic spectrum allocation
	based on reinforcement learning algorithms in cognitive radio networks,''
	{\em Artificial Intelligence Review}, vol.~51, no.~3, pp.~493--506, 2019.
	
	\bibitem{caminero2019adversarial}
	G.~Caminero, M.~Lopez-Martin, and B.~Carro, ``Adversarial environment
	reinforcement learning algorithm for intrusion detection,'' {\em Computer
		Networks}, vol.~159, pp.~96--109, 2019.
	
	\bibitem{li2011smart}
	X.~Li, R.~Lu, X.~Liang, X.~Shen, J.~Chen, and X.~Lin, ``Smart community: an
	internet of things application,'' {\em IEEE Communications magazine},
	vol.~49, no.~11, pp.~68--75, 2011.
	
	\bibitem{andrea2015internet}
	I.~Andrea, C.~Chrysostomou, and G.~Hadjichristofi, ``Internet of things:
	Security vulnerabilities and challenges,'' in {\em 2015 IEEE Symposium on
		Computers and Communication (ISCC)}, pp.~180--187, IEEE, 2015.
	
	\bibitem{roman2013features}
	R.~Roman, J.~Zhou, and J.~Lopez, ``On the features and challenges of security
	and privacy in distributed internet of things,'' {\em Computer Networks},
	vol.~57, no.~10, pp.~2266--2279, 2013.
	
	\bibitem{alaba2017internet}
	F.~A. Alaba, M.~Othman, I.~A.~T. Hashem, and F.~Alotaibi, ``Internet of things
	security: A survey,'' {\em Journal of Network and Computer Applications},
	vol.~88, pp.~10--28, 2017.
	
	\bibitem{alanazi2015resilience}
	S.~Alanazi, J.~Al-Muhtadi, A.~Derhab, K.~Saleem, A.~N. AlRomi, H.~S.
	Alholaibah, and J.~J. Rodrigues, ``On resilience of wireless mesh routing
	protocol against dos attacks in iot-based ambient assisted living
	applications,'' in {\em 2015 17th International Conference on E-health
		Networking, Application \& Services (HealthCom)}, pp.~205--210, IEEE, 2015.
	
	\bibitem{antonakakis2017understanding}
	M.~Antonakakis, T.~April, M.~Bailey, M.~Bernhard, E.~Bursztein, J.~Cochran,
	Z.~Durumeric, J.~A. Halderman, L.~Invernizzi, M.~Kallitsis, {\em et~al.},
	``Understanding the mirai botnet,'' in {\em 26th $\{$USENIX$\}$ Security
		Symposium ($\{$USENIX$\}$ Security 17)}, pp.~1093--1110, 2017.
	
	\bibitem{woolf}
	N.~Woolf, ``Ddos attack that disrupted internet was largest of its kind in
	history, experts say.''
	
	\bibitem{inbook}
	I.~Romdhani, R.~Abdmeziem, and D.~Tandjaoui, {\em Architecting the Internet of
		Things: State of the Art}.
	\newblock 07 2015.
	
	\bibitem{wu2010research}
	M.~Wu, T.-J. Lu, F.-Y. Ling, J.~Sun, and H.-Y. Du, ``Research on the
	architecture of internet of things,'' in {\em 2010 3rd International
		Conference on Advanced Computer Theory and Engineering (ICACTE)}, vol.~5,
	pp.~V5--484, IEEE, 2010.
	
	\bibitem{wang2016internet}
	P.~Wang, S.~Chaudhry, L.~Li, S.~Li, T.~Tryfonas, and H.~Li, ``The internet of
	things: a security point of view,'' {\em Internet Research}, 2016.
	
	\bibitem{zhong2015study}
	C.-L. Zhong, Z.~Zhu, and R.-G. Huang, ``Study on the iot architecture and
	gateway technology,'' in {\em 2015 14th International Symposium on
		Distributed Computing and Applications for Business Engineering and Science
		(DCABES)}, pp.~196--199, IEEE, 2015.
	
	\bibitem{finkenzeller2009known}
	K.~Finkenzeller, ``Known attacks on rfid systems, possible countermeasures and
	upcoming standardisation activities,'' in {\em 5th European Workshop on RFID
		Systems and Technologies}, pp.~1--31, 2009.
	
	\bibitem{mitrokotsa2010classification}
	A.~Mitrokotsa, M.~R. Rieback, and A.~S. Tanenbaum, ``Classification of rfid
	attacks,'' {\em Gen}, vol.~15693, no.~14443, p.~14, 2010.
	
	\bibitem{chien2007security}
	H.-Y. Chien and C.-W. Huang, ``Security of ultra-lightweight rfid
	authentication protocols and its improvements,'' {\em ACM SIGOPS Operating
		Systems Review}, vol.~41, no.~4, pp.~83--86, 2007.
	
	\bibitem{sonar2014survey}
	K.~Sonar and H.~Upadhyay, ``A survey: Ddos attack on internet of things,'' {\em
		International Journal of Engineering Research and Development}, vol.~10,
	no.~11, pp.~58--63, 2014.
	
	\bibitem{deng2005defending}
	J.~Deng, R.~Han, and S.~Mishra, ``Defending against path-based dos attacks in
	wireless sensor networks,'' in {\em Proceedings of the 3rd ACM workshop on
		Security of ad hoc and sensor networks}, pp.~89--96, 2005.
	
	\bibitem{malialis2013multiagent}
	K.~Malialis and D.~Kudenko, ``Multiagent router throttling: Decentralized
	coordinated response against ddos attacks,'' in {\em Twenty-Fifth IAAI
		Conference}, 2013.
	
	\bibitem{malialis2015distributed}
	K.~Malialis and D.~Kudenko, ``Distributed response to network intrusions using
	multiagent reinforcement learning,'' {\em Engineering Applications of
		Artificial Intelligence}, vol.~41, pp.~270--284, 2015.
	
	\bibitem{yau2005defending}
	D.~K. Yau, J.~C. Lui, F.~Liang, and Y.~Yam, ``Defending against distributed
	denial-of-service attacks with max-min fair server-centric router
	throttles,'' {\em IEEE/ACM Transactions on Networking}, vol.~13, no.~1,
	pp.~29--42, 2005.
	
	\bibitem{hu2014survey}
	F.~Hu, Q.~Hao, and K.~Bao, ``A survey on software-defined network and openflow:
	From concept to implementation,'' {\em IEEE Communications Surveys \&
		Tutorials}, vol.~16, no.~4, pp.~2181--2206, 2014.
	
	\bibitem{fang2013loss}
	S.~Fang, Y.~Yu, C.~H. Foh, and K.~M.~M. Aung, ``A loss-free multipathing
	solution for data center network using software-defined networking
	approach,'' {\em IEEE transactions on magnetics}, vol.~49, no.~6,
	pp.~2723--2730, 2013.
	
	\bibitem{liu2018deep}
	Y.~Liu, M.~Dong, K.~Ota, J.~Li, and J.~Wu, ``Deep reinforcement learning based
	smart mitigation of ddos flooding in software-defined networks,'' in {\em
		2018 IEEE 23rd International Workshop on Computer Aided Modeling and Design
		of Communication Links and Networks (CAMAD)}, pp.~1--6, IEEE, 2018.
	
	\bibitem{8276173}
	D.~G. {Bhoyar} and U.~{Yadav}, ``Review of jamming attack using game theory,''
	in {\em 2017 International Conference on Innovations in Information, Embedded
		and Communication Systems (ICIIECS)}, pp.~1--4, 2017.
	
	\bibitem{grover2014jamming}
	K.~Grover, A.~Lim, and Q.~Yang, ``Jamming and anti-jamming techniques in
	wireless networks: a survey,'' {\em International Journal of Ad Hoc and
		Ubiquitous Computing}, vol.~17, no.~4, pp.~197--215, 2014.
	
	\bibitem{weber2010towards}
	S.~G. Weber, L.~Martucci, S.~Ries, and M.~M{\"u}hlh{\"a}user, ``Towards
	trustworthy identity and access management for the future internet,'' in {\em
		4th International Workshop on Trustworthy Internet of People, Things \&
		Services}, 2010.
	
	\bibitem{wu2011anti}
	Y.~Wu, B.~Wang, K.~R. Liu, and T.~C. Clancy, ``Anti-jamming games in
	multi-channel cognitive radio networks,'' {\em IEEE journal on selected areas
		in communications}, vol.~30, no.~1, pp.~4--15, 2011.
	
	\bibitem{el2014power}
	R.~El-Bardan, S.~Brahma, and P.~K. Varshney, ``Power control with jammer
	location uncertainty: A game theoretic perspective,'' in {\em 2014 48th
		Annual Conference on Information Sciences and Systems (CISS)}, pp.~1--6,
	IEEE, 2014.
	
	\bibitem{cagalj2006wormhole}
	M.~Cagalj, S.~Capkun, and J.-P. Hubaux, ``Wormhole-based antijamming techniques
	in sensor networks,'' {\em IEEE transactions on Mobile Computing}, vol.~6,
	no.~1, pp.~100--114, 2006.
	
	\bibitem{wang2011anti}
	B.~Wang, Y.~Wu, K.~R. Liu, and T.~C. Clancy, ``An anti-jamming stochastic game
	for cognitive radio networks,'' {\em IEEE journal on selected areas in
		communications}, vol.~29, no.~4, pp.~877--889, 2011.
	
	\bibitem{7841922}
	N.~{Namvar}, W.~{Saad}, N.~{Bahadori}, and B.~{Kelley}, ``Jamming in the
	internet of things: A game-theoretic perspective,'' in {\em 2016 IEEE Global
		Communications Conference (GLOBECOM)}, pp.~1--6, 2016.
	
	\bibitem{6523802}
	R.~D. {Pietro} and G.~{Oligeri}, ``Jamming mitigation in cognitive radio
	networks,'' {\em IEEE Network}, vol.~27, no.~3, pp.~10--15, 2013.
	
	\bibitem{8320276}
	H.~A. {Bany Salameh}, S.~{Almajali}, M.~{Ayyash}, and H.~{Elgala}, ``Spectrum
	assignment in cognitive radio networks for internet-of-things delay-sensitive
	applications under jamming attacks,'' {\em IEEE Internet of Things Journal},
	vol.~5, no.~3, pp.~1904--1913, 2018.
	
	\bibitem{heo2017dodge}
	J.~Heo, J.-J. Kim, S.~Bahk, and J.~Paek, ``Dodge-jam: Anti-jamming technique
	for low-power and lossy wireless networks,'' in {\em 2017 14th Annual IEEE
		International Conference on Sensing, Communication, and Networking (SECON)},
	pp.~1--9, IEEE, 2017.
	
	\bibitem{kim2015cognitive}
	S.~Kim, ``Cognitive radio anti-jamming scheme for security provisioning iot
	communications.,'' {\em KSII Transactions on Internet \& Information
		Systems}, vol.~9, no.~10, 2015.
	
	\bibitem{rawat2015securing}
	D.~B. Rawat and M.~Song, ``Securing space communication systems against
	reactive cognitive jammer,'' in {\em 2015 IEEE Wireless Communications and
		Networking Conference (WCNC)}, pp.~1428--1433, IEEE, 2015.
	
	\bibitem{djuraev2017channel}
	S.~Djuraev, J.-G. Choi, K.-S. Sohn, and S.~Y. Nam, ``Channel hopping scheme to
	mitigate jamming attacks in wireless lans,'' {\em EURASIP Journal on Wireless
		Communications and Networking}, vol.~2017, no.~1, p.~11, 2017.
	
	\bibitem{becker2014dynamic}
	J.~Becker, ``Dynamic beamforming optimization for anti-jamming and hardware
	fault recovery,'' 2014.
	
	\bibitem{8417695}
	Y.~{Chen}, Y.~{Li}, D.~{Xu}, and L.~{Xiao}, ``Dqn-based power control for iot
	transmission against jamming,'' in {\em 2018 IEEE 87th Vehicular Technology
		Conference (VTC Spring)}, pp.~1--5, 2018.
	
	\bibitem{7952524}
	G.~{Han}, L.~{Xiao}, and H.~V. {Poor}, ``Two-dimensional anti-jamming
	communication based on deep reinforcement learning,'' in {\em 2017 IEEE
		International Conference on Acoustics, Speech and Signal Processing
		(ICASSP)}, pp.~2087--2091, 2017.
	
	\bibitem{khan2017cognitive}
	A.~A. Khan, M.~H. Rehmani, and A.~Rachedi, ``Cognitive-radio-based internet of
	things: Applications, architectures, spectrum related functionalities, and
	future research directions,'' {\em IEEE wireless communications}, vol.~24,
	no.~3, pp.~17--25, 2017.
	
	\bibitem{xiao2018anti}
	L.~Xiao, X.~Wan, W.~Su, Y.~Tang, {\em et~al.}, ``Anti-jamming underwater
	transmission with mobility and learning,'' {\em IEEE Communications Letters},
	vol.~22, no.~3, pp.~542--545, 2018.
	
	\bibitem{8314744}
	X.~{Liu}, Y.~{Xu}, L.~{Jia}, Q.~{Wu}, and A.~{Anpalagan}, ``Anti-jamming
	communications using spectrum waterfall: A deep reinforcement learning
	approach,'' {\em IEEE Communications Letters}, vol.~22, no.~5, pp.~998--1001,
	2018.
	
	\bibitem{li2019performance}
	Y.~Li, X.~Wang, D.~Liu, Q.~Guo, X.~Liu, J.~Zhang, and Y.~Xu, ``On the
	performance of deep reinforcement learning-based anti-jamming method
	confronting intelligent jammer,'' {\em Applied Sciences}, vol.~9, no.~7,
	p.~1361, 2019.
	
	\bibitem{7636793}
	S.~{Machuzak} and S.~K. {Jayaweera}, ``Reinforcement learning based
	anti-jamming with wideband autonomous cognitive radios,'' in {\em 2016
		IEEE/CIC International Conference on Communications in China (ICCC)},
	pp.~1--5, 2016.
	
	\bibitem{bkassiny2012wideband}
	M.~Bkassiny, S.~K. Jayaweera, Y.~Li, and K.~A. Avery, ``Wideband spectrum
	sensing and non-parametric signal classification for autonomous self-learning
	cognitive radios,'' {\em IEEE Transactions on Wireless Communications},
	vol.~11, no.~7, pp.~2596--2605, 2012.
	
	\bibitem{7925694}
	M.~A. {Aref}, S.~K. {Jayaweera}, and S.~{Machuzak}, ``Multi-agent reinforcement
	learning based cognitive anti-jamming,'' in {\em 2017 IEEE Wireless
		Communications and Networking Conference (WCNC)}, pp.~1--6, 2017.
	
	\bibitem{babu2010comprehensive}
	P.~R. Babu, D.~L. Bhaskari, and C.~Satyanarayana, ``A comprehensive analysis of
	spoofing,'' {\em International Journal of Advanced Computer Science and
		Applications}, vol.~1, no.~6, pp.~157--62, 2010.
	
	\bibitem{4358709}
	Z.~{Duan}, X.~{Yuan}, and J.~{Chandrashekar}, ``Controlling ip spoofing through
	interdomain packet filters,'' {\em IEEE Transactions on Dependable and Secure
		Computing}, vol.~5, no.~1, pp.~22--36, 2008.
	
	\bibitem{whalen2001introduction}
	S.~Whalen, ``An introduction to arp spoofing,'' {\em Node99 [Online Document],
		April}, 2001.
	
	\bibitem{7804660}
	M.~{Nawir}, A.~{Amir}, N.~{Yaakob}, and O.~B. {Lynn}, ``Internet of things
	(iot): Taxonomy of security attacks,'' in {\em 2016 3rd International
		Conference on Electronic Design (ICED)}, pp.~321--326, 2016.
	
	\bibitem{liu2011robust}
	F.~J. Liu, X.~Wang, and H.~Tang, ``Robust physical layer authentication using
	inherent properties of channel impulse response,'' in {\em 2011-MILCOM 2011
		Military Communications Conference}, pp.~538--542, IEEE, 2011.
	
	\bibitem{liu2017active}
	J.~Liu, L.~Xiao, G.~Liu, and Y.~Zhao, ``Active authentication with
	reinforcement learning based on ambient radio signals,'' {\em Multimedia
		Tools and Applications}, vol.~76, no.~3, pp.~3979--3998, 2017.
	
	\bibitem{7398138}
	L.~{Xiao}, Y.~{Li}, G.~{Han}, G.~{Liu}, and W.~{Zhuang}, ``Phy-layer spoofing
	detection with reinforcement learning in wireless networks,'' {\em IEEE
		Transactions on Vehicular Technology}, vol.~65, no.~12, pp.~10037--10047,
	2016.
	
	\bibitem{xiao2019learning}
	L.~Xiao, W.~Zhuang, S.~Zhou, and C.~Chen, ``Learning-based rogue edge detection
	in vanets with ambient radio signals,'' in {\em Learning-based VANET
		Communication and Security Techniques}, pp.~13--47, Springer, 2019.
	
	\bibitem{bezzo2018predicting}
	N.~Bezzo, ``Predicting malicious intention in cps under cyber-attack,'' in {\em
		2018 ACM/IEEE 9th International Conference on Cyber-Physical Systems
		(ICCPS)}, pp.~351--352, IEEE, 2018.
	
	\bibitem{8046382}
	E.~{Yel}, T.~X. {Lin}, and N.~{Bezzo}, ``Reachability-based self-triggered
	scheduling and replanning of uav operations,'' in {\em 2017 NASA/ESA
		Conference on Adaptive Hardware and Systems (AHS)}, pp.~221--228, 2017.
	
	\bibitem{ramachandran2007bayesian}
	D.~Ramachandran and E.~Amir, ``Bayesian inverse reinforcement learning.,'' in
	{\em IJCAI}, vol.~7, pp.~2586--2591, 2007.
	
	\bibitem{elnaggar2018irl}
	M.~Elnaggar and N.~Bezzo, ``An irl approach for cyber-physical attack intention
	prediction and recovery,'' in {\em 2018 Annual American Control Conference
		(ACC)}, pp.~222--227, IEEE, 2018.
	
	\bibitem{pillitteri2014guidelines}
	V.~Y. Pillitteri and T.~L. Brewer, ``Guidelines for smart grid cybersecurity,''
	tech. rep., 2014.
	
	\bibitem{wang2013survey}
	D.~Wang, X.~Guan, T.~Liu, Y.~Gu, Y.~Sun, and Y.~Liu, ``A survey on bad data
	injection attack in smart grid,'' in {\em 2013 IEEE PES Asia-Pacific Power
		and Energy Engineering Conference (APPEEC)}, pp.~1--6, IEEE, 2013.
	
	\bibitem{zhu2014sequential}
	Y.~Zhu, J.~Yan, Y.~Tang, Y.~Sun, and H.~He, ``The sequential attack against
	power grid networks,'' in {\em 2014 IEEE International Conference on
		Communications (ICC)}, pp.~616--621, IEEE, 2014.
	
	\bibitem{yan2016q}
	J.~Yan, H.~He, X.~Zhong, and Y.~Tang, ``Q-learning-based vulnerability analysis
	of smart grid against sequential topology attacks,'' {\em IEEE Transactions
		on Information Forensics and Security}, vol.~12, no.~1, pp.~200--210, 2016.
	
	\bibitem{chen2018evaluation}
	Y.~Chen, S.~Huang, F.~Liu, Z.~Wang, and X.~Sun, ``Evaluation of reinforcement
	learning-based false data injection attack to automatic voltage control,''
	{\em IEEE Transactions on Smart Grid}, vol.~10, no.~2, pp.~2158--2169, 2018.
	
	\bibitem{8514804}
	M.~N. {Kurt}, O.~{Ogundijo}, C.~{Li}, and X.~{Wang}, ``Online cyber-attack
	detection in smart grid: A reinforcement learning approach,'' {\em IEEE
		Transactions on Smart Grid}, vol.~10, no.~5, pp.~5174--5185, 2019.
	
	\bibitem{loch1998using}
	J.~Loch and S.~P. Singh, ``Using eligibility traces to find the best memoryless
	policy in partially observable markov decision processes.,'' in {\em ICML},
	pp.~323--331, 1998.
	
	\bibitem{8603817}
	Z.~{Ni} and S.~{Paul}, ``A multistage game in smart grid security: A
	reinforcement learning solution,'' {\em IEEE Transactions on Neural Networks
		and Learning Systems}, vol.~30, no.~9, pp.~2684--2695, 2019.
	
	\bibitem{karagiannis2011vehicular}
	G.~Karagiannis, O.~Altintas, E.~Ekici, G.~Heijenk, B.~Jarupan, K.~Lin, and
	T.~Weil, ``Vehicular networking: A survey and tutorial on requirements,
	architectures, challenges, standards and solutions,'' {\em IEEE
		communications surveys \& tutorials}, vol.~13, no.~4, pp.~584--616, 2011.
	
	\bibitem{8855737}
	N.~{Alsaffar}, H.~{Ali}, and W.~{Elmedany}, ``Smart transportation system: A
	review of security and privacy issues,'' in {\em 2018 International
		Conference on Innovation and Intelligence for Informatics, Computing, and
		Technologies (3ICT)}, pp.~1--4, 2018.
	
	\bibitem{javed2016security}
	M.~A. Javed, E.~Ben~Hamida, and W.~Znaidi, ``Security in intelligent transport
	systems for smart cities: From theory to practice,'' {\em Sensors}, vol.~16,
	no.~6, p.~879, 2016.
	
	\bibitem{8231220}
	L.~{Xiao}, C.~{Xie}, M.~{Min}, and W.~{Zhuang}, ``User-centric view of unmanned
	aerial vehicle transmission against smart attacks,'' {\em IEEE Transactions
		on Vehicular Technology}, vol.~67, no.~4, pp.~3420--3430, 2018.
	
	\bibitem{8246580}
	L.~{Xiao}, X.~{Lu}, D.~{Xu}, Y.~{Tang}, L.~{Wang}, and W.~{Zhuang}, ``Uav relay
	in vanets against smart jamming with reinforcement learning,'' {\em IEEE
		Transactions on Vehicular Technology}, vol.~67, no.~5, pp.~4087--4097, 2018.
	
	\bibitem{lu2017anti}
	X.~Lu, D.~Xu, L.~Xiao, L.~Wang, and W.~Zhuang, ``Anti-jamming communication
	game for uav-aided vanets,'' in {\em GLOBECOM 2017-2017 IEEE Global
		Communications Conference}, pp.~1--6, IEEE, 2017.
	
	\bibitem{8761101}
	S.~{Feng} and S.~{Haykin}, ``Anti-jamming v2v communication in an integrated
	uav-cav network with hybrid attackers,'' in {\em ICC 2019 - 2019 IEEE
		International Conference on Communications (ICC)}, pp.~1--6, 2019.
	
	\bibitem{weng2018bandit}
	L.~Weng, ``The multi-armed bandit problem and its solutions,'' {\em
		lilianweng.github.io/lil-log}, 2018.
	
	\bibitem{auer2002finite}
	P.~Auer, N.~Cesa-Bianchi, and P.~Fischer, ``Finite-time analysis of the
	multiarmed bandit problem,'' {\em Machine learning}, vol.~47, no.~2-3,
	pp.~235--256, 2002.
	
	\bibitem{lazaric2008reinforcement}
	A.~Lazaric, M.~Restelli, and A.~Bonarini, ``Reinforcement learning in
	continuous action spaces through sequential monte carlo methods,'' in {\em
		Advances in neural information processing systems}, pp.~833--840, 2008.
	
	\bibitem{kulkarni2016hierarchical}
	T.~D. Kulkarni, K.~Narasimhan, A.~Saeedi, and J.~Tenenbaum, ``Hierarchical deep
	reinforcement learning: Integrating temporal abstraction and intrinsic
	motivation,'' in {\em Advances in neural information processing systems},
	pp.~3675--3683, 2016.
	\balance
\end{thebibliography}

\end{document}